\documentclass{article}

\usepackage{PRIMEarxiv}

\usepackage[utf8]{inputenc} 
\usepackage[T1]{fontenc}    
\usepackage{hyperref}       
\usepackage{url}            
\usepackage{booktabs}       
\usepackage{amsfonts}       
\usepackage{nicefrac}       
\usepackage{microtype}      
\usepackage{lipsum}
\usepackage{fancyhdr}       
\usepackage{graphicx}       
\graphicspath{{media/}}     

\title{A Survey of Traversability Estimation for
Mobile Robots
}

\author{
  Christos Sevastopoulos\\
  Department of Computer Science and Computer Engineering \\
  University of Texas at Arlington \\
  Arlington, TX, USA\\
  christos.sevastopoulos@mavs.uta.edu \\
   \And
  Stasinos Konstantopoulos \\
  Institute of Informatics and Telecommunications \\
 NCSR Demokritos \\
  Agia Paraskevi,Greece\\
  \texttt{konstant@iit.demokritos.gr} \\
}

\begin{document}
\maketitle

\begin{abstract}

Traversability illustrates the difficulty of driving through a specific region and encompasses
the suitability of the terrain for traverse based on its physical properties, such as slope and roughness, surface
condition, etc. In this survey we highlight the merits and limitations of all the major steps in the evolution
of traversability estimation techniques, covering both non-trainable and machine-learning methods, leading
up to the recent proliferation of deep learning literature. We discuss how the nascence of Deep Learning has
created an opportunity for radical improvement in traversability estimation. Finally, we discuss how self-
supervised learning can help satisfy deep methods’ increased need for (challenging to acquire and label)
large-scale datasets.

\end{abstract}


\section{Introduction}
In leading-edge mobile robotics research, a wide array of outdoor navigation applications such as planetary exploration, military operations, agricultural tasks etc. entails the necessity of adapting to the conditions encountered and, in particular, to address the challenges imposed by the terrain's contextual complexity. For every mobile robot, it is of indispensable importance to be able to identify its surroundings and translate the information perceived by its sensors to a meaningful volume of required knowledge. Subsequently, it will obtain the capacity to determine whether it can navigate in a safe while efficient manner. A vital part of autonomous navigation implies that the perceived structure of the environment has to be precisely illustrated in order to ensure whether a specific region can be traversed or not. Uneven terrains characterized by dense vegetation,foliage and potential presence of obstacles create a fundamental need to choose carefully between an proprioceptive or a exteroceptive sensor modality. It is apparent that the selection of the sensors used has to be in accordance with each application's prerequisites and the robot's structural design.

Traversability illustrates the difficulty of driving through a specific region and encompasses the suitability of the terrain for traverse based on its physical properties, such as slope and roughness, surface condition, etc. \cite{seraji1999traversability} as well as the mechanical characteristics and capabilities of the robot. Furthermore, it might also establish a bedrock for path planning algorithms  since it is inevitably incorporated in certain terrain indices such as terrain roughness, terrain inclination etc. which are of central importance when considering the most feasible path \cite{ishigami2011path}. 
Although at an initial stage \cite{papadakis2013terrain}, traversability estimation was recognized as a binary classification problem, currently it can be viewed through the prism of multiple classes categorization with respect to the levels of traverse facilitation. Being able to evaluate terrains' traversability is a constitutional step towards designing a perception system \cite{langer1994behavior} for such rough and rugged terrains while processing voluminous data acquired by different sensory techniques.

 Deep Vision expands the traversability estimation field, as it facilitates the detection of features that conventional geometry-based approaches do not have access to; that is estimating compliance from visual information. Strictly speaking, although, from a geometric point of view,  the presence of obstacles might insinuate that the paths appear to be non-traversable, a robot might still be able to penetrate a compliant obstruction such as grass,foliage etc. Thereupon, it would be vital to determine whether Deep Learning Vision techniques can enrich the environment's perception with semantic environment information on top of geometric information and hence introduces a further ability to navigate such environments. Various studies have expounded the importance of traversability analysis as a fundamental step towards motion planning. Papadakis \cite{papadakis2013terrain} presents a wide-angle review of how the multi-sensor acquisition of input data, incorporating laser/stereo/color information etc. juxtaposed with the accurate representation of vehicle models interacting with the terrain, can lead to meaningful inference of traversability estimation for structured and unstructured environments. In \cite{kostavelis2015semantic}, various methods of extracting semantic mapping information are being discussed along with their potential applications in the field of mobile robotics, including traversability assessment.

\begin{figure}[tb]
    \centering
    \includegraphics[width=9cm]{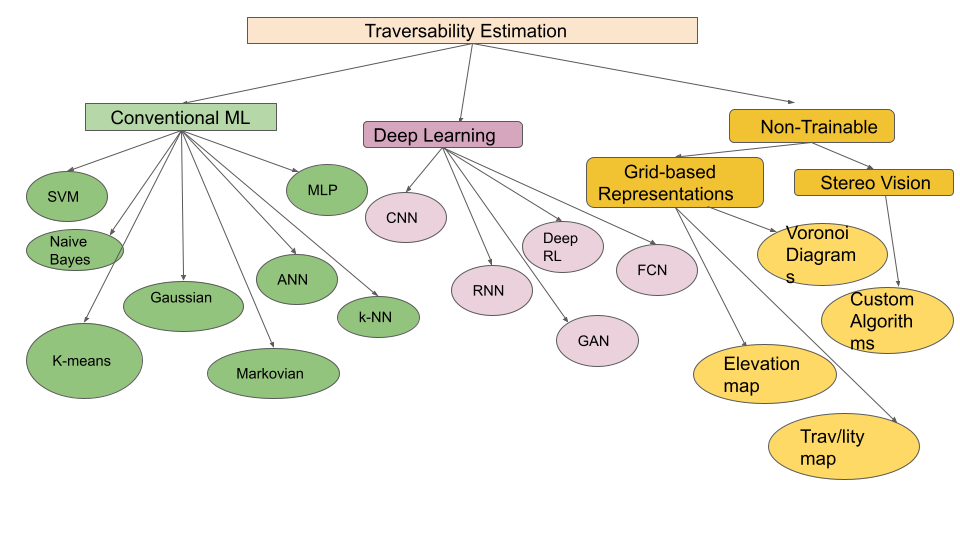}
    \caption{Overview of the different techniques discussed in this survey}
    \label{fig:overview}
\end{figure}

\begin{figure}[tb]
    \centering
    \includegraphics[width=9cm]{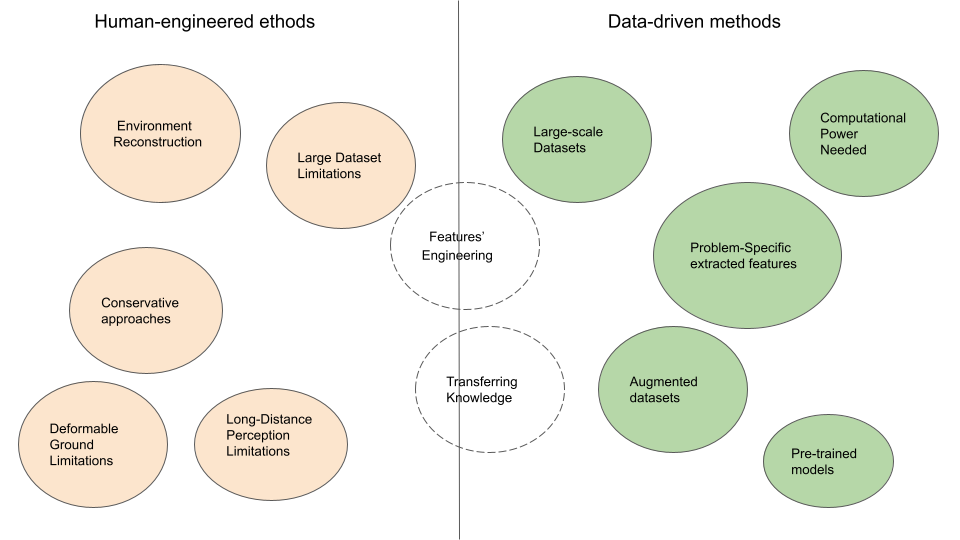}
    \caption{Facilitations and limitations between conventional vision and machine learning/deep learning techniques}
    \label{fig:beforeafter}
\end{figure}

Having the opportunity to examine the traversability analysis notion from a variety of research angles, this survey paper delineates the contemporary advances in traversability estimation through the prism of Deep Learning techniques as well as a juxtaposition along conventional machine learning and non-trainable methods (Figure \ref{fig:overview}). Inferring terrain's traversability from geometric information can frequently bump into limitations as a consequence of the problems' high dimensionality while meaningful information is extracted from image data. Therefore, due to the intricacy that complex terrains portray, this article aims
to illustrate how the nascence of Deep Learning, and especially Self-Supervised Learning (SSL),  exhibits advances in traversability estimation scenarios where the crucial need for accurate labelling of large-scale data can be ponderous when performed by a human expert. Figure \ref{fig:beforeafter} presents some highlights, in terms of advantages and limitations that human-engineered and data-driven methods exhibit.

Summarizing, we aim to present the traversability estimation problem through the following viewpoints:
\begin{itemize}
    \item highlight important elements of certain geometric and conventional machine learning techniques
    \item present an overview of state of the art deep learning techniques
    \item discuss the challenges that arise in different learning scenarios 
    \item point out why SSL has the qualities to be the most promising direction
\end{itemize}

\section{Non-Trainable Methods}
\subsection{Grid-based Representation}
In order to construct an appropriate discretization of the environment that encloses the required knowledge about the places that have been traversed, sensory information is transformed using techniques such as occupancy grids, digital elevation maps, and traversability maps. Principally, the elevation map consists of a two-dimensional regular grid, where each cell stores a height value and variance. While the robot explores the environment and collects new information, this map is continuously updated and traditionally, terrain traversability would refer to grid cell traversability computation. With regards to elevation maps, they can be built by using on-board sensing such as lidars and IMUs and by exploiting the geometry of the adjacency grid \cite{pan2019gpu}, traversability can be determined and thereupon, initiate the ground for successful motion planning.

Similarly to the elevation map, the traversability map uses the regular grid representation. A 2D grid-based approach, using fused stereo and visual data 
that segments the environment into equally sized spatial cells \cite{sock2016probabilistic}, while the principal difference against the occupancy grid-map is that, in this case,  each cell illustrates the traversability rather than the occupancy of the space examined. 
A local traversability value is assigned to each cell of the elevation map and by computing the traversability map,  the traversability of a certain robot pose can be interpreted. Traversability maps' direct and intuitive nature relies heavily on the fact that they can be built with respect to each sensors acquired findings. For instance, Wermelinger et al.  \cite{wermelinger2016navigation} construct the traversability map by incorporating three fundamental terrain characteristics: slope, terrain roughness and step height. In \cite{fan2021step} the map is represented as a collection of terrain properties (e.g., height, risk) over a uniform grid and the estimation of  traversability shall encircle a number of potential threats, such as collisions, step size, tip-over, contact loss, slippage and uncertainty in sensors, along with localization errors towards creating a planning framework by addressing the problem as an Model Predictive Control (MPC) problem.

Properly identifying and calculating the values of these aforementioned topographical characteristics, along with the robot's mechanical capacity \cite{li2019rugged}, an enriched traversability elevation map can be constructed. Consequently, it facilitates dynamic exploration tasks since it is going to dictate the exact location on which e.g. a walking robot can successfully land a valid step by adjusting its position and orientation accordingly \cite{kim2020vision}, \cite{loc2011improving}. 

 Alternatively, obtaining the traversability map can be an intermediate step towards calculating robot's control commands (steering, velocity). 
 In particular, in \cite{ye2007navigating} a terrain map is built by laser sensor findings and is then converted to a traversability map by assigning a Traversability Index (TI) value to each cell in the terrain map. Ultimately, through the utilization of an one dimensional histogram (Traversability Field Histogram), the robot can efficiently navigate to its target location. Martin et al. \cite{martin2013building} highlight through their experiments  how, by exploiting the use of  onboard sensors (GPS, accelerometers, gyroscopes etc.) four traversability metrics (power consumption, longitudinal slip, lateral slip and vehicle orientation) can efficiently generate traversability costmaps.

%

\subsection {Conventional Computer Vision}

Early Computer Vision (CV) techniques on traversability estimation relied on making predictions based on the output that obstacle detection algorithms yielded. The system presented in \cite{ulrich2000appearance}, 
 utilizes monocular color vision data. It focuses on the appearance of individual pixels and local visual attributes, such as intensity, color,  edges, and  texture. The quintessence of the method presented consists of detecting pixels different in appearance than the ground and classifying them as obstacles i.e. Meaning that the detection of any discrepancy between a single pixel  that  differs in appearance  from  the  ground is classified  as an obstacle. Similarly, Huertas et al. \cite{huertas2005stereo} 
elaborate on the differentiation of the boundaries that tree trunks can have against the background as decided by an edge detection algorithm. 

Stereo imagery, along with color,hue,texture information can be practical in efforts of building a multi-algorithm approach \cite{rankin2005evaluation} that is independently detecting numerous obstacles  of governing characteristics such as tree trunks, water, excessive slope etc. by taking into account the terrain's attributes. Using  stereo modelling and outliers detection,Bajracharya et al. \cite{bajracharya2013high} build a uniform terrain mapping system that incorporates information on elevation,slope,roughness along with categorization of positive and negative obstacles. 
In specific, they focus on distinguishing thin structures with respect to the large amount of depth singularities and rich textural information involved, such as grass or sparse bushes, that cannot pose a genuine threat to the robot's safety.

Based on terrain features, such as slope and roughness, Castejon et al., \cite{castejon2005traversable} 
estimate the traversability characteristics, by exploiting the power of Voronoi Diagrams to model the XY dimensions of the outdoor environment as well as creating a qualitative representation that defines the traversability model. The latter provides useful geometrical information for constructing the Digital Elevation Map that eventually contributes in discretizing the workspace and isolate the cells that offer traversability information. As a means to execute a traversability analysis in complex catacomb-like environments, Bogoslavskyi et al.  \cite{bogoslavskyi2013efficient}
  perform experiments on a mobile robot solely collecting input from depth images drawn by a Kinect-style sensor. The use a sequential way of extracting the traversability interpretation starting from local traversability as a result of a single depth image and afterwards, proceed with integrating all the single-image traversability estimates, into a local traversability map. The way to ensure the efficacy of their method is to perform a pixel by pixel comparison between the traversability estimates and the custom-made structures with known 3D geometry. During certain evaluation trials, it is found that the traversable regions were dependent on a specific steering of the robot and thus their method is facing serious limitations.

\section{Conventional Machine Learning}

Early endeavors in incorporating sensory input with machine learning techniques determining the presence of an obstacle  or taking an action based on the environment's structure are described by \cite{dima2004classifier} 
and \cite{pomerleau1992progress}  respectively. In specific, Dima et al.\cite{dima2004classifier} present a framework governed by the use of multiple sensors such as lasers, camera and infrared imagery that contribute towards detecting humans, negative obstacles and terrain's traversability. This is achieved through merging the strengths of three different classifiers (AdaBoost, stacked generalization, experts)  on the manually collected data. Pomerleau \cite{pomerleau1992progress} explores how the performance of a neural network-based autonomous driving improves via human demonstration. An Artificial Neural Network (ANN) receiving camera input is shown to produce more accurate levels of the output i.e steering of an autonomous vehicle by employing domain-specific knowledge during its training process.

Given the fact that unstructured environments often embody large levels of uncertainty, regression methods are engaged in the endeavor of traversability estimation. Ho et al. \cite{ho2013traversability} employ the Gaussian Process (GP) Regression technique on the way to predict a planetary rover’s attitude and configuration angles by learning the vehicle response on unstructured terrain from experience. Hence, the estimation of traversability relies heavily on the formulation of the GP regression problem. Using exteroceptive data as the training input, they aim towards a direct calculation of traversability by adopting  an architecture for estimating the kernel function in order to monitor the evolution of vehicle states and propagation of uncertainty. Subsequently, GP provides a continuous representation of the terrain, and accurate estimation of traversability in areas with little or no exteroceptive data. They conclude that, combining exteroceptive with proprioceptive learning can yield more complete and more accurate traversability maps. Extending the Gaussian Process regression model, Oliveira et al.  \cite{oliveira2020bayesian} implement an Uncertain-Inputs GP model that provides the usefulness of scrutinizing localization and execution noise as an effort of modeling terrain's roughness using vibration data as an input.

Support Vector Machine (SVM) \cite{mohri2018foundations}, a kernel-based method has been in established usage for tasks involving classification, regression, novelty detection etc.  due to its ability to make classification decisions for new input vectors rather than yield probabilistic outputs \cite{bishop2006pattern}. Predominantly, it provides general solutions for maximizing the margin between two particular classes. In the direction of detecting road traversability,  Bellone et al. \cite{bellone2017learning} carefully select a feature set generated through normal vector analysis. The hypothesis made implies that by using a normalized descriptor, that is enriched with both geometric and color data, augments the generalization of the space descriptor. Veritably, their proposed descriptor,through the use of an SVM with 4 different kernels, manages to portray higher level of efficiency than certain standard descriptors and  it can subsequently detect road traversability from  point  clouds  acquired  in  outdoor environments. Zhou et al. \cite{zhou2012self} use the AdaBoost algorithm associated with Fuzzy SVM  for feature selection on a 3D point cloud for the ground surface, in order to create a self-supervised visual learning for terrain surface detection in forest environments. As a means to train the classifier, a triangulated irregular network (TIN) is employed to model the ground plane and extract training points from the 3D point cloud dataset.

Alternatively, an attempt to exploit the features of combined color or depth descriptors and thus generating a textural descriptor is described by \cite{narvaez2018terrain}.  These textural, along with the color, features were accepted as training and testing inputs in the SVM classifier for performing terrain classification on areas covered with sand, grass, pavement, gravel and litterfall. Another research attempt in which textural features take a leading role on the way to derive terrain's classification is described by \cite{kingry2018vision}. In particular, key features are extracted from captured visual-spectrum  images and the use of an Artificial Neural Network (ANN) facilitates the way to identify terrain types of grass, concrete, asphalt, mulch, gravel, and dirt.

\begin{figure}[tb]
    \centering
    \includegraphics[width=9cm]{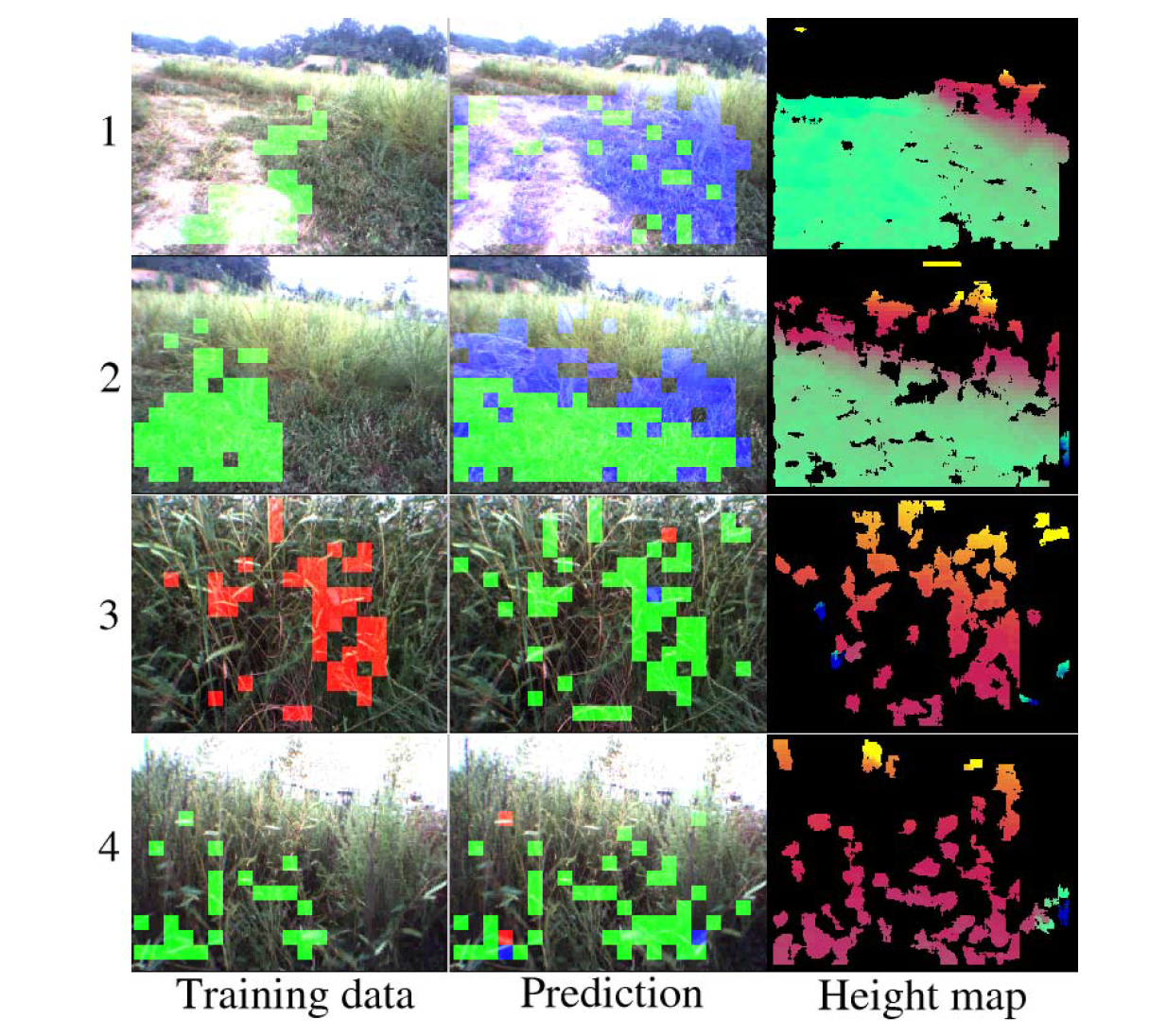}
    \caption{ For the training/prediction data, different colors correspond to  traversable (green),non-traversable(red) and uncertain(blue) patches. The resulting heightmap  represents the ascending levels of elevation in the "blue to green to red to yellow" respectively \cite{kim2006traversability} }
    \label{fig:kimtrav}
\end{figure}

An instance of incrementally training an unsupervised learning scheme, where the classifier's ultimate goal is to provide predictions about the visible terrain's traversability is delineated in \cite{kim2006traversability}. Autonomously collected by stereo, labeled visual features corresponding to traversable or nontraversable examples   are fed to an online classifier learning algorithm that its operation aligns with the axes of 1) identifying and collecting appearance feature vectors from training examples, and 2) classifying newly collected image patches  based on the learned models.
The efficiency of the running mechanism articulates that labeled data gathered within a certain time window ought to be used for training as a means to fruitfully map the image patches input to the terrain map cells. Subsequently, the desired output will highlight areas classified as traversable,non-traversable or unknown (Figure \ref{fig:kimtrav}). After performing experiments, it was shown that the traversability that was on-line learnt could enable the robot's access to a goal location despite being surrounded by tall grass of high density while a conventional planner, which function counts exclusively on estimating the cost map from the elevation map  computed through stereo ranging, was proven to be unsuccessful as a result of lack of representational power of the feature space.

An endeavor embodying the integration of geometry and traversability en route to generating a terrain's assessment is presented in \cite{happold2006enhancing}. In this sense, a multi-layer perceptron (MLP) with one hidden layer is trained with stereo images,in a supervised manner, on eight-dimensional geometric vectors depicting features such as slope,density, height,vertical distance etc. The LAGR platform collects stereo pairs and then the human expert labels each explored cell as low, intermediate, high, or lethal with respect to the difficulty encountered while traversing this particular cell. After gathering a total of 4000 labeled cells, it is shown that height and slope were the most important features in terms of determining terrain's traversability. Subsequently, by integrating the classification from geometry features with color information, a cost map was generated for path planning purposes.

\subsection{Probabilistic} 
Thrun et al.\cite{thrun2002probabilistic}  make a statement of fundamental gist that '\textit{As robots are moving away from factory floors into increasingly unstructured environments,the  ability to cope with uncertainty is critical for building successful robots'}. Navigation in outdoor complex environments encapsulates the need to handle the uncertainty risen as an amalgam of different factors such as sensors' noise and error, robot's mechanical limitations and most importantly the environment's unpredictable nature which renders its modelling as a quite challenging task. A plethora of approaches that can represent uncertainty using probabilistic distributions and modelling has been introduced for deriving terrain traversability. 
In the work of Ollis et al. \cite{ollis2007bayesian} 
the robot learns from human demonstration to calculate terrain costs, which indicate the probability of an obstacle's presence. Through the combination of using Bayesian estimates and geometric information collected by stereo vision, the final terrain costs follow a certain distribution and thus it can be determined whether the path is traversable by articulating that those cells with higher values of features ought to be less traversable. Using the human interference in a similar manner but excluding any presuming correlation between the features and the traversability, in  \cite{suger2015traversability} 
training data is generated throughout the safe journey that the human operator drives the robot throughout. Thereupon, the notion of learning is addressed with the application of the Positive Naive Bayes (NB) classifier 
that estimates the frequencies of observed features by finding the parameters of the probability distribution for the traversability. 
As being examined within the aforementioned parts of this survey, the formulation of the traversability map
acts as a powerful tool in representing the path that the robot can safely pursue.
Furthermore, in order to augment the accuracy in detection, Sock et al. \cite{sock2016probabilistic} 
fuse, using the Bayes’ formula that combines input from independent sources to estimate a single entity, the traversability maps obtained by a visual camera and a LIDAR. The resulting map is modelled as a Markov Random Field and the cells are being independently updated. 

Another regularly implemented technique which aims to autonomously improve traversability  estimation capabilities in unknown terrains is the use of a self-learning framework   
where 3D information corresponding to a densely vegetated terrain is extracted from the point cloud and is afterwards fed, through the form of geometric features, to a geometry-based classifier \cite{reina2012towards}. The main rationale en route to estimate the ground's traversability implies that the geometry classifier supervises a second color-based classifier and hence an iterative process that the system is retrained while new labelled data enriches the representation of the ground's model which is constructed with the use of Gaussian Mixture Models (GMM). Leveraging the advantages that a self-learning framework offers reciprocally with the use of \textit{superpixels} as visual primitives, Kim et al. \cite{kim2007traversability} 
employ vision sensing to estimate the traversability of the  terrain based on its  appearance instead of using the geometric stereo vision information. Their superpixel-based approach which produces higher levels of accuracy on image region classification, involving features containing color and texture information in RGB, computes traversability using Bayes' rule along with a modified k-nearest neighborhood (k-NN) algorithm. As a way to distinguish between known and unknown regions, i.e. frontiers of the traversability map obtained by laser scans during autonomous exploration, Tang et al.\cite{tang2019autonomous} 
make use of the reachability map that reduces the traversability map's dimension. By enforcing the k-means clustering method on the frontier candidates, along with the use of the A* Algorithm for finding the optimal paths, the cells on the grip map are being labelled as reachable,dangerous or unknown. Furthermore, defining the boundaries, upper and lower, of the terrain map Fankhauser et al. \cite{fankhauser2018probabilistic} 
propose a mapping approach using proprioceptive sensing (kinematic and inertial measurements) 
relying on the current pose of the robot that is being constantly updated as well as the noise and uncertainty of the sensor and roll,pitch angles respectively. The gist of their method is built upon creating a robust robot-centric elevation map that generates its data through the uncertainty derived from the robot's incremental motion in the form of mathematical equations. Although their experiments using legged robotic hardware managed to apprehend and make use of the environment's uncertainty, their implementation seems to be facing limitations that the authors manage to address for maps of larger size along with localization singularities due to their platform-specific method. Similarly, another platform-specific effort that the proposed robot-centric mapping system aiming to derive traversability using laser-based 3D SLAM is illustrated by Droeschel et al. \cite{droeschel2017continuous}
Using proprioceptive sensing (IMU and local odometry) along with rotating laser scanner measurements that in the surface element representation are going to be interpreted as Gaussian Mixture Model observations, a pose graph is assembled by the maps of all the adjacent key poses in the direction of successfully computing the robot's localization. By performing graph optimization, local dense 3D maps are constructed and integrated to a global one that yields information for the robot's real time pose and can ultimately provide traversability costs for rough terrain navigation for each map cell.

A conjunctional viewpoint of both the terrain's geometry and kinematic configuration of the robot is explored by the work of  \cite{belter2016adaptive} 
in which a prediction-based terrain traversability assessment method relying heavily on the RRT algorithm is presented. By creating a reference map for prediction, the algorithm determines the path the walking robot follows upon acting on a reference map created prior experiments on rough terrains. Traversability is decided upon evaluating footholds,feet trajectories and other constraints among the cells of the map.

\section {Deep Learning}

\begin{figure*}[t]
    \centering
    \includegraphics[width=\textwidth]{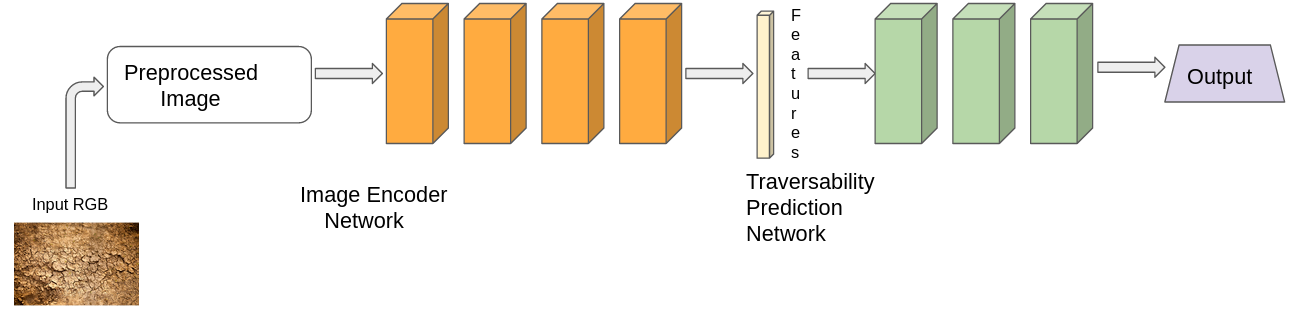}
    \caption{An example pipeline describing the process of identifying the meaningful features through a series of convolutions before determining the traversability output through a subsequent network.}
    \label{fig:pipeline}
\end{figure*}

While conventional machine learning techniques can face serious restrictions in terms of being able to process the collected data in their initial state, Deep Learning methods offer the prospect of creating better representations and thus leading to better understanding without onerous engineering struggles. A Deep Learning asset that can go beyond conventional machine learning methods in traversability estimation scenarios is that it offers the prospect of creating implicit relationships among data. Traversability estimation has been examined from the various perspectives of unsupervised,semi-supervised,supervised and self-supervised learning. Contemporary methods, are often associated with models being trained in an end-to-end supervised fashion, as a means to simplifying the training process. Equivalently, unsupervised and semi-supervised methods that use pre-trained models' features shortly before training on a supervised dataset for a specific downstream task have been a sharing a large amount of popularity too. A common approach in deep vision traversability estimation techniques pinpointed by the latest research efforts, is to process an input RGB image through a series of pre-processing techniques and convolutional layers of a self-supervised network, before the meaningful features are fed to a traversability prediction network,i.e a classifier (Figure \ref{fig:pipeline}).

\subsection{Supervised}

Supervised learning has been a prevailing tool in traversability computation for rough terrains due to its applicability in predicting and classifying terrains and regions that correspond to a specific class. Especially, after obtaining a training set, input sensory data needs to be mapped to a specific target value frequently concerning whether the terrain is traversable or not. 

After generating a significant volume of training data, in \cite{chavez2018learning}  
it is investigated whether a traversability classifier learnt from synthetic heightmaps performs well on real heightmaps. Towards this direction, the traversability estimation is addressed as a heightmap classification problem and a comparison to determine whether a patch is traversable is performed between a standard computer-vision feature extracting technique and a CNN one.
As far as the deep learning architecture built, it consists of adjacent convolutional and max-pooling layers followed by a Fully Convolutional (FC) layer with 2 output neurons. It is reported that the Convolutional Neural Network (CNN) estimator outperforms the feature-based approaches both on synthetic and real-world heightmaps. Besides that, it is noted that the training on simulation could successfully get transfered to traversability maps reflecting unseen real-world terrains.

In the interest of performing long-range terrain segmentation using RGB stereo images in outdoor environments,  Zhang et al. \cite{zhang2018long} design an end-to-end training deep CNN architecture aiming to augment the network's generalization efficiency. An encoder-decoder scheme using input feature and reference maps (calculated by the disparity images) is used. In particular, the encoder includes five layer ensembles consisting of convolutional layers followed by batch normalization ones right before the RELU activation function and the pooling layers. The decoder uses upsampling layers that perform a deconvolution operation on the input features maps before the latter being processed by the aforementioned ensemble. It is reported that the introduction of the reference maps at the 1st, 3rd, and 5th ensemble layers provided a good balance between the segmentation guidance and the noise suppression. Subsequently, the output of the decoder is fed into a multi-class softmax classifier to generate predicted labels while performing pixel-wise classification. For their experiments, the used six hand-labeled image datasets containing RGB images and the disparity maps drawn from areas of dirt, foliage, natural obstacles (trees and dense shrubs), mulch etc. Subsequently, each pixel is classified into one of three terrain classes of traversable, nontraversable, and unknown regions respectively.

\begin{table*}[t]

\renewcommand*{\arraystretch}{1.2}
\caption{Overview of papers using Supervised learning } \label{Supervisedtable}
\begin{center}
\begin{tabular}{|c|c|c|c|c|}
  \hline
 Reference & Input &  NumOfData  & Architecture & Platform-specific  \\
  \hline\hline
 Pfeiffer et al., 2017 \cite{pfeiffer2017perception} & 2D Laser,Position   & 4.3M    & CNN & Yes\\ \hline
 Palazzo et al., 2020 \cite{palazzo2020domain} & RGB &  5k   & CNN  & No \\ \hline
 Tai et al., 2017 \cite{tai2017autonomous} & RGB-D &  1104   & CNN  &    No\\ \hline

 Zhang et al., 2018b \cite{zhang2018robot} & RGB & 700    & CNN  &    No\\ \hline
Kingry et al., 2018 \cite{kingry2018vision} & RGB  &  15k segments  & ANN  &    No\\ \hline
Chavez-Garcia et al., 2018 \cite{chavez2018learning} & Heightmap Patch &  65k  & CNN  &    No\\ \hline

Bellone et al., 2017  \cite{bellone2017learning}  & RGB Point Cloud & 5k    &  SVM  &    No\\ \hline
  Rothrock et al., 2016 \cite{rothrock2016spoc}
  & RGB &  700   &  FCNN &    Yes\\ \hline
  Sock et al., 2016 \cite{sock2016probabilistic} & 3D Laser , RGB &  215k   & SVM  &    No\\ \hline

 Narváez et al., 2018  \cite{narvaez2018terrain} & Depth,IR, \par Color   &820 & SVM& No  \\ \hline

Ollis et al., 2007 \cite{ollis2007bayesian} & RGB & Unspec  & Bayesian & No  \\ \hline

Ho et al., 2013 \cite{ho2013traversability} & RGB-D \par Point Clouds & 215k & GP Regression & Yes\\ \hline

Dima et al., 2004 \cite{dima2004classifier}  & IR,Color & 28k & Stacked Generalization& No \\
&&& Committee of experts, AdaBoost &  \\ \hline

Olivera et al., 2020 \cite{oliveira2020bayesian} & IMU & Unspec  &  GP Regression &  No \\ \hline
Gao et al., 2021 \cite{gao2021fine} & RGB,IMU,\par Laser &12k & CNN &No \\ \hline
\multicolumn{3}{p{300pt}}{Unspec = Unspecified
}\\
\end{tabular}
\end{center}
\end{table*}

An assessment of terrain traversability for performing autonomous classification of Martian terrains is explored in \cite{rothrock2016spoc} 
where a framework named Soil Property and Object Classifier (SPOC) that provides pixel-by-pixel image classification into one of 17 terrain classes is constructed upon a CNN. In a supervised fashion, human experts append the imagery dataset with new classes while the Mars Rover explores different sorts of terrains. A fully differentiable and trained for 6 hours end-to-end FC network ,including multiple stages of filtering, various CNN layer dimensions (64,128,256,512) and downsampling, is used as the understructure before an upsampled penultimate layer classifies the input raw image (orbital or ground). The output of the classification acts as the input to the cost function of the optimal route planner for landing site traversability analysis and also for building a robust slip prediction model.

A human-like inspired system trained in a supervised way involving the fusion of CNNs with decision-making process was explored in the work of Tai et al. \cite{tai2017autonomous} 
where the network's output generates the control commands for the robot to explore an indoor environment. In terms of the deep learning model used, an input depth map is fed to a three-times repeated assemblage of convolution/activation/pooling plus an FC layer of five nodes each one corresponding to robot's the possible states (going forward, 2x turning left, 2x turning right). Using a Turtlebot and a Kinect sensor, the training set comprised of indoor depth information while the ground-truth was defined by a human operator and the robot's decision-making actions demonstrated similarities to human-inspired intelligence.


In \cite{b50}, through the use of a global motion planner, the robot learns a navigation policy in a supervised manner with the use of a CNN by fusing  preprocessed collected laser data  
with the relative goal position 
The CNN architecture presented encompasses two residual building blocks including shortcut connections that can address training complexity. Moreover it is shown that, they train and test two CNN models in simulation (with a small differentiation in their dimensions), 
before expanding to a real platform that traverses an area with obstacles such as tables,chairs etc. Their results indicate that their model was not only able to learn the desired navigation strategies but also to transfer the knowledge among different unseen environments. However, some impediments occur with the rise of the environment's complexity since it is stated that the CNN is not able to act as global path planner.

In contrast with methods that exhibit pure adherence to frames' binary classification as traversable or not, Palazzo et al.  \cite{palazzo2020domain}
design a supervised model that can analyze multiple traversability routes through the medium of the encoder-decoder architecture. Notably, while the problem is examined as a regression one, their aim is to estimate and predict the traversability costs of various routes even on scenarios that no labels are provided. Using collected RGB images as inputs, the utilized architecture consists of a FC network module for feature extraction, followed by two layers; a convolutional and a fully-connected respectively. The bottom line of their method lies on training a model to predict correct traversability scores on the source dataset, while carrying out unsupervised domain adaptation on the target data.

Table \ref{Supervisedtable} provides an overview of papers using Supervised Learning along with the input used, the total number of data used in the experiment, the architecture used and also describes whether the method used is only suited to the platform used in the experiments. The same structure will be applied to  Tables \ref{SSLupervisedtable}, \ref{Hybridtable} \ref{nonSUPERVISEDtable}
of the following sections.

\subsection{Self-Supervised}

Self-supervised learning (SSL) is a form of supervised learning that human intervention, in terms of labeling, is not necessitated. In specific, the agent investigates a partition of unlabeled data, interprets it, and then, by developing a reliable representation, it is able to produce the labels missing and thus develop a sturdy perspective about the remaining part while automatically creating a labeled dataset. A key aspect of SSL that renders it as the contemporary most promising direction towards traversability estimation in unknown environments, is the ability to establish larger proportions of data efficiency in deep learning models that aim,  as a consequence of reduced demand, for hand-labeled training data. Subsequently, it battles against the pure reliance on extensive amounts of data, and it is proven to be highly beneficial especially for scenarios that involve updated data collections for different tasks as described by\cite{sekar2020planning}.

One of the first endeavors in determining long-range traversability using  short-range data and self-supervision is described in  \cite{brooks2007self}. In pursuance of training a vision classifier for a four-wheeled  
rover in a Martian-like rough terrain, the authors  
present two self-supervised approaches for local and distant terrain classification respectively. Short range data input, acquired both from vision and vibration sensors, creates a "local training" framework fusing texture,color and geometry information for all the encountered classes i.e rock,sand, grass. Using the short range training, the second approach for "remote training" employs stereo processing to identify the distance to patches in the image and, by position estimation, to identify when the rover has driven over a particular patch of the terrain. Consequently, long-range data was classified with respect to the classes having previously been identified for the "local" scenario. Collecting data while the rover traverses the terrain and setting a threshold for the data points collected for the visual classifier, training was realized with an SVM classifier, and by fusing the class likelihoods of the color, visual texture, and geometric sensing modes using Naive Bayes, terrain classification is performed.

Another work, interpolating a long-range vision classifier trained in self-supervision is portrayed by Hadsell et al. \cite{hadsell2009learning}. 
The classifier's output allow successful detection of trees,obstacles etc., by having the horizon as its perspective , and thus determining the traversability of the input large image patch patches. With regards to the features extractors involved,  a total of four approaches  working in an interleaved fashion is presented, each one being trained with either labeled or unlabeled offline data. The efficiency reported in their results is produced by the use of a multilayer CNN that was initialized with deep belief net training, consisting of two convolutional and max-pooling between them layers, which was responsible for independently pretraining each layer in both unsupervised and supervised manners.

Significant amounts of research attention due to recent advances in the field of Deep Learning is placed on the novelty detection problem for indoor/outdoor robot navigation. A distinguishable work has been presented by Richter et al. \cite{richter2017safe} in which the novelty detection scenario is addressed through the utilization of autoencoders irrespective of the extent of appropriate training that the robot has received. In detail, the robot repeatedly gathers training data, labels it in a self-supervised manner and are fed to a conventional feedforward neural network consisting of three hidden layers,followed by sigmoid activation functions and a softmax output layer that predicts collisions or not. On the other end, the autoencoder, comprising of three hidden sigmoid layers and a sigmoid output layer, reconstructs similar inputs and determines whether the new images bare enough resemblance to those of the training data. In the case of detecting something novel in the environment, the robot decides towards a safe behavior. In any other case, it can just augment the array of familiar environment types. Nonetheless, instances of misclassification of novelty in images might occur due to inadequate training of the collision predictor. The authors address this incident by matching the size and architecture of the hidden layers of the two networks.

Analogously, Wellhausen et al. \cite{wellhausen2020safe} adopt similar reasoning in their study in which the robot, collects RGB-D images in a self-supervised way, and gets trained on 10000 training image patches corresponding to traversable areas. With regards to the autoencoders architecture, the encoder uses three consecutive blocks, each containing a convolutional layer of kernel 5, followed by a ReLU function. The first two blocks are followed by a MaxPooling layer, whereas the final block is followed by an additional final convolution with kernel size 1. On the other side, the decode comprises of the same architecture but it uses nearest-neighbor upscaling layers instead of MaxPooling.

A deep neural network architecture consisting of multiple stream as a means to address the learning of features of different modalities, is presented by 
Valada et al. in \cite{valada2017adapnet} where using 
the convoluted mixture of deep experts (CMoDE) fusion technique, they provide a semantic segmentation technique that relies heavily on the use of ResNet \cite{he2016deep} and dilated convolutions. They perform platform-specific  experiments on a forested environment incorporating scenarios of detrimental nature for the robot such as low-lighting, motion blur,occlusions etc. Regardless of the absence of prior map, their tests showed that the robot could successfully traverse the trail by using on the fly semantic segmentation.

An alternative approach adopting self-supervised learning is presented by Shah et al. \cite{shah2021ving} where traversability is determined upon learning a traversability function 'T' which describes whether any controller can successfully navigate among collected observations (RGB images). This aforementioned navigation policy shall also take into account the environment's physics, rather than pure geometry, in order to decide which objects, tall grass for instance, are traversable or not. Subsequently, the ultimate goal of the policy is to successfully predict the estimated number of time steps required by a controller P to navigate from one observation to another. Both T and P architectures encompass the use of a MobileNet encoder \cite{kingma2014adam} followed by three dense layers that, using supervised learning, they project 
1024-dimensional latents from the collected images to 50 class labels and 3 waypoint outputs corresponding to the relative pose between the collected observations for the cases of T and P respectively. 
These two learnt functions along with past gathered experience gained by arbitrarily chosen trajectories is unified to a system 'VinG' that governs a goal-oriented behaviour that solely relies on offline experience.
En route to introducing a path planning framework that leverages learning 
of terrain's visual representations used in conjunction with unlabelled human navigation
demonstrations, Sikand et al. \cite{sikand2021visual} present a work that aims  to create a mapping from the representation space to the terrain costs that the robot encounters while traversing a specific terrain. Visual data is collected during a self-supervised procedure in which the robot traverses a setting including sidewalks,trees,and terrains of grass,dirt etc. With a small human demonstration intervention, the datasets created, comprise of triplets of image patches corresponding to an anchor patch, a \textit{similar} patch that portrays the same image as the anchor but from a different viewpoint and \textit{dissimilar} one that includes information of patches that the human avoids choose and are far from the embedding space. As a means to form the visual representation, a CNN with two convolutional layers followed by three FC layers is used.

\begin{figure*}[]
    \centering
    \includegraphics[width=\textwidth]{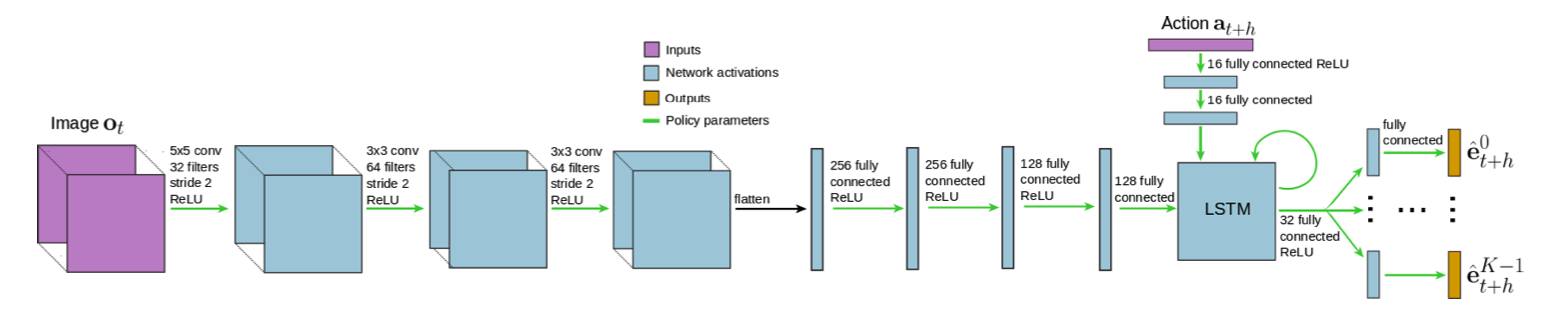}
    \caption{An example of a deep neural network predictive model \cite{kahn2021badgr} receiving an RGB image as an input and processing it throughout a series of convolutional,FCNs and LSTM layers. The outputs are subsequently fed to a motion planner that determines the action needed to avoid a bumpy terrain or a potential collision}
    \label{fig:badgrarch}
\end{figure*}

A conjointment of supervision between an unsupervised acoustic proprioceptive classifier that self-supervises an exteroceptive visual one is explored by Zurn et al. \cite{zurn2020self} 
. The robot equipped with both a stereo camera and a microphone, traverses various complex terrains and collects visual (terrain patches) and audio (vehicle-terrain interaction) data respectively.
This is achieved by associating the visual features of a ground's patch right in front of the robot with the auditory features that correspond to the area that the robot is traversing. Projecting the camera images into a birds-eye-view perspective, they act as weakly labeled training data for the semantic segmentation network trained in an unsupervised manner. By teleoperating a rubber-wheeled robot, they collect visual (24 thousand images) and auditory data (4 hours of video) of five different terrains: Asphalt, Grass, Cobblestone, Parking Lot, and Gravel. In regard to the architecture used, the authors implement an  encoder and decoder architecture for the audio data and  the MobileNet V2  model pretrained on the ImageNet dataset for the visual feature extraction network. Moreover, for the semantic segmentation of the terrains, they adopt the AdapNet++ network with an EfficientNet backbone. Their self-supervised exteroceptive semantic segmentation model achieved a comparable performance to supervised learning with manually labeled data.

An automated self-supervised learning method and the corresponding prediction of navigation-relevant terrain properties is presented in 
\cite{wellhausen2019should}  
In specific, they conduct their experiments using the ANYmal robot \cite{hutter2016anymal} by measuring the interaction,during locomotion, between the robot's sensorized feet and the terrain and then projecting the robot's footholds into camera images. Using this foothold projection system, they annotate semantic classes in the images by assigning a semantic label to each time step in the sequence while human annotation is only implicated in observing and marking the possible transitions between terrain types with a concrete time stamp and the terrain's type. The possible terrain types involved are asphalt, gravel path, grass, dirt, and sand. A learnt, while walking, terrain property, named 'ground reaction score',  provides an alternative way to generate the  footholds' labels in a self-supervised way. For the semantic segmentation purposes of this study, the authors use a CNN which architecture is based on ERFNet \cite{romera2017erfnet}.
By teleoperating the robot through different environments, they collect a dataset of 70000 training and 15000 validation images respectively. Ultimately, in addition to the application of their approach to a legged robot like ANYmal, it is mentioned that it could be deployed to other types of ground robots as well.

\begin{table*}[t]

\renewcommand*{\arraystretch}{1.2}
\caption{Overview of papers using Self-Supervised learning } \label{SSLupervisedtable}
\begin{center}
\begin{tabular}{|c|c|c|c|c|}
  \hline
 Reference & Input &  NumOfData & Architecture & Platform-specific \\
  \hline\hline
   Sikand et al., 2021 \cite{sikand2021visual}  &  Image Patches  & Unspec  & CNN  &No  \\ \hline
   Shah et al., 2021 \cite{shah2021ving} &  RGB &  Unspec  & CNN encoder  &  No  \\ \hline
 Kahn et al., 2021 \cite{kahn2021badgr} &  RGB & 720k datapoints & CNN,\par LSTM  & Yes \\\hline
  Wellhausen et al., 2019 \cite{wellhausen2019should} & RGB &  86k   & CNN  &    Yes\\ \hline
 Valada et al., 2017 \cite{valada2017adapnet} & RGB & 5000   &  Mixture of Experts&  No \\ \hline

 Richter and Roy, 2017 \cite{richter2017safe} & RGB-D &  10k   & Auto \par Encoders&    No\\ \hline


Zhou et al., 2012 \cite{zhou2012self} & Laser Point Cloud  & 650k+ L &SVM & No  \\ \hline
Reina and Milella , 2012 \cite{reina2012towards}& Stereo & Unspec & Gaussian Mixture Model & Yes\\ \hline

Hadsell et al., 2009 \cite{hadsell2009learning} & Stereo & 10k& CNN & Yes  \\ \hline
 Brooks and Iagnemma, 2007 \cite{brooks2007self}& Stereo & Unspec & SVM,\par Naive Bayes& Yes  \\ \hline
Kim et al., 2007 \cite{kim2007traversability} & Stereo & Unspec& KNN & No  \\ \hline

\multicolumn{3}{p{300pt}}{Unspec = Unspecified
}\\

\end{tabular}
\end{center}
\end{table*}

\subsection{Unsupervised \& Semi-Supervised}

Semi-supervised and unsupervised learning provide auspicious ground for focusing on the essential segments  rather than precise pixel-wise classification that require labelling of the training images.

One preceding method is described by Shneier et al. \cite{shneier2008learning} using range and color information, build unsupervised models solely derived from the geometry and appearance of the scene. These models are learnt using clusters of neighboring data, extracted from the same physical region and can enclose an estimate of the traversability cost. However, due to the fact that the features learnt within the models are self-sufficient, not requiring any range data, causes the association between the region models and the traversability estimation to be uncertain for distant regions.

In recent research, a fundamental aim of the use of unsupervised learning in traversability estimation problems is to to learn a particular set of features that can be transfered to the network that will subsequently be trained on a specific downstream task(such as an image classification one) which determines traversable areas. Transfer learning approaches normally require a large-volume training dataset (e.g., Pascal Voc, Kitti, Imagenet) to train, and by using the pre-trained weights of a model, a classification task can yield higher levels of accuracy and abstraction.

An instance of a deep-learning-based architecture, considering the unsupervised problem as supervised, that aims to train a generative model, \cite{goodfellow2020generative} is served by the use of Generative Adversarial Networks. GANs have been in wide research use in computer vision tasks over the span of the previous decade and can automatically train a generative model while using both a generative and a discriminative model.

Exploiting the notion of transfer learning,
Hirose et al. \cite{hirose2017go}  
train a model with positive examples automatically spawn by an on-board fisheye camera within a time window of 7.2 hours. With regards to the architecture presented, it consists of the conventional ensemble of the two adversarial modules i.e a generator and a discriminator which are both designed as standard CNNs which are trained in a merely supervised fashion since it is noted that only a slight annotation provided drastic improvement on the performance. On top of that, the generator receives the latent vector z as an input which has been previously generated by an additional network, the inverse generator that is being trained simultaneously along with the aforementioned generator and discriminator. In order to enhance the levels of accuracy of the unsupervised method, an additional Fully Convolutional (FC) layer is trained in a supervised way to classify the scenes as "GO" or "NOT GO". This linear classifier is using  the GAN knowledge extracted by three specific GAN features.  In order to test the network's effectiveness, a saliency map is used to provide the meaningful segments of the images which correspond to the right and left side of the input images as they indicate the presence of a wall or a corridor and thus portray a fundamental role in determining indoor traversability. Ultimately, it is stated that the method presented could be an imperative tool towards the effort of building cost maps.

Extending their work in performing GAN predictions for indoor traversability scenarios Hirose et al. \cite{hirose2018gonet} 
introduce the GONet framework that uses the same aforementioned idea of semi-supervised learning that incorporates a small amount of negative training data that can be proven to be more advantageous than solely including positive data, in improving traversability estimation.  Indicatively, the GONet architecture consists of two models, one responsible for extracting features from positive automatically labeled examples of traversable areas extracted by a fisheye camera mounted on a robot and the second which performs the final classification after having been trained on poth positive and negative examples. Additionally, in order to exploit the temporal nature of the collected data, a Long Short Term Memoy (LSTM) unit that captures the temporal dependencies in the data is added, creating a new separate model named GONet+T, and the output it yields is further fed to an FC layer responsible for subsequently predicting the traversability probability. Strengthening the performance of GONet, a second extension named GONet+TS that is trained identically as the GONet+T and  addresses the limitations in prediction owing to environment's structural  captured in stereo images  is introduced.  
Performing indoor experiments with the TurtleBot2 platform and using the saliency map,  despite the effectiveness presented in all methods performed,  GONet+T and GONet+TS  highlighted  smoother predictions in indoors traversability estimation than GONet due to the inclusion of the LSTM layer.


Utilizing GONet's application with a dynamic-scene view synthesis method \cite{hirose2019vunet} , named 'VUNet', the authors present a unified system that can single out the traversable areas in the robot's vicinity.
 VUNet is the result of combining two supplementary networks SNET and DNet that have the ability to model static and dynamic transformations based on robot's actions. SNET is responsible for predicting static (S) and DNET for predicting dynamic (D) changes in the  parts of the environment due to robot motion respectively. SNEt 's architecture is based on the encoder-decoder scheme, and ought to the fact that the sampling procedure reuses original pixels of the input image, sharper images are generated. On the other hand, DNet is built upon a conditional adversarial network architecture.
 As a consequence, due to the bifold character of that synthesized approach, both static (e.g., walls, windows, stairs) and dynamic (e.g., humans) components of the environment can be predicted from different camera poses in future time steps. In order to estimate future traversability, two applications based on assisted teleoperation are introduced  i.e. early obstacle detection (moving pedestrians)  and multi-path future traversability estimation. 
As inputs to VUNEt,  the last two acquired images and a virtual navigation command, i.e., a linear and angular velocity are used. During the first experiment, VUNet predicts the motion of the human in the image and informs the teleoperator using warning signals and emergency stop commands. With regards to the second application, the system is able to generate virtual velocities for five different paths around the robot, and hence by predicting the images using scene view method, it can compute the traversability for each of the paths.

Using GONet and VUNet as a solid baseline, an 8-convolutional layer architecture named PoliNet \cite{hirose2019deep} is trained to learn the Model Predictive Control-policy (MPC) 
for performing safe visual navigation of a mobile robot with mere human supervision. Concretely, by combining VUNet-360 ,a variant of VUNet that uses input from a 360 camera, with the aforementioned traversability estimation network GONet, PoliNet can produce the velocity commands necessary for the robot to successfully follow a visual path in a safe manner. The control police tries to enforce minimization of the difference between an image taken from the 360 camera at time \textit{t} and the next sub-goal image in the trajectory. Hence, the control policy is responsible for finding the appropriate location in a way that the current image looks similar to the one of the sub-goal's. PoliNet is trained offline before getting transferred to the online setup. Data was collected both in simulation and in the real world and with regards to the real data, the robot was tele-operated and gathered 10 a half hours of 360 camera RGB images. Although, their experiments displayed high amounts of their method's robustness, there were instances in which the robot was not able to circumvent large obstacles, mainly due to the fact that traversability was only considered as a soft constraint in the optimization problem.

Training GANs occasionally suffers from an array of reasons such as catastrophic forgetting \cite{thanh2020catastrophic} as well as difficulties in convergence, mode collapse and instability, due to design-related issues i.e  network architecture, appropriate selection of objective function etc. \cite{saxena2021generative}. For the cases of gathering data in an unsupervised manner or with scarce labels, such as an autonomous visual data collection by a mobile robot, recent advances in self-supervised contrastive learning offer the advantage of optimizing the learning capabilities of the designed model or operating in conjunction with the semi-supervised learning approach that is tailored to the downstream task that is examined. For instance, in \cite{goh2022mars}, using the popular approach of SimCLR \cite{chen2020simple}, they perform  a martian terrain segmentation analysis with limited data corresponding to classes such as oil, bedrock, sand,big rock, rover and background. Using supervised contrastive learning, Gao et al. \cite{gao2021fine} manually label a set of anchor patches in their effort to efficiently create a feature representation that is able to distinguish different traversability regions.
In \cite{shah2022viking}, the output of an heuristic model trained on teleoperated prior data using the contrastive  InfoNCE loss function \cite{van2018representation}, is combined with the output of a local traversability model towards successful path planning.

\begin{table*}[t]

\renewcommand*{\arraystretch}{1.2}
\caption{Overview of papers using Hybrid learning } \label{Hybridtable}
\begin{center}
\begin{tabular}{|c|c|c|c|c|c|}
  \hline
 Reference & Input &  NumOfData &  Labels & Architecture & Platform-specific \\
  \hline\hline

Goh et al., 2022\cite{goh2022mars}& Stereo & 17k & SSL, \par Super & CNN &No \\\hline
Shah and Levine,  \cite{shah2022viking} & RGB & Unspec  & SSL, Super &CNN & No\\
2022  & GPS & && &    \\\hline

Zürn et al., 2020 \cite{zurn2020self} & Stereo,  & 25k & SSL, & CNN & No\\
& Audio & &Unsuper & Autoencoder& \\\hline
Sekar et al., 2020 \cite{sekar2020planning}  & RGB, & 1000steps & SSL,RL & CNN  & No \\ 
& &/episode & & RNN,MLP & \\\hline
Hirose et al., 2017 \cite{hirose2017go} & RGB & 78k & Super, \par Unsuper &  GAN, \par FCL &  No \\\hline
Tai and Liu, 2016 \cite{tai2016towards} & RGB-D & 32images/iteration & Super, RL & CNN, FCN & No \\
& & 4k iterations &  & &\\\hline
Happold et al., 2006  \cite{happold2006enhancing} & Stereo  & 4k & Super, Unsuper  & MLP & No  \\ \hline

\multicolumn{3}{p{300pt}}{Unspec = Unspecified
}\\
\end{tabular}
\end{center}
\end{table*}

\subsection{Deep Reinforcement Learning}

As described by Sutton \cite{sutton2018reinforcement},  \textit{in uncharted territory-where one would expect learning to be most beneficial-an agent must be able to learn from his own experience}
Reinforcement learning (RL) enables a robot to autonomously discover an optimal behavior through trial-and error interactions\cite{kober2013reinforcement}. What differentiates the field of robotics in terms of applying reinforcement learning to, is the amount of challenges encompassed. One major arduousness is the high dimensionality and complexity of the states involved as well as the adversity in performing a complete and noise-free observation of the true state. What is more, another considerable difficulty that RL is facing in Robotics derives from the fact that interactions between a mechanical system and its environment can harm the platform or any humans involved. However, RL can be proven to be an effective arrow in the quiver when the robot is navigating through complex and dynamic environments \cite{polydoros2017survey}. On top of that, conventional RL algorithms fused with Deep Learning, can handle many practical problems, where the incorporated states of the Markov Decision Process exhibit high levels of dimensionality and thus optimal policies are easier to be learnt.

\begin{figure*}[tb]
    \centering
    \includegraphics[width=\textwidth]{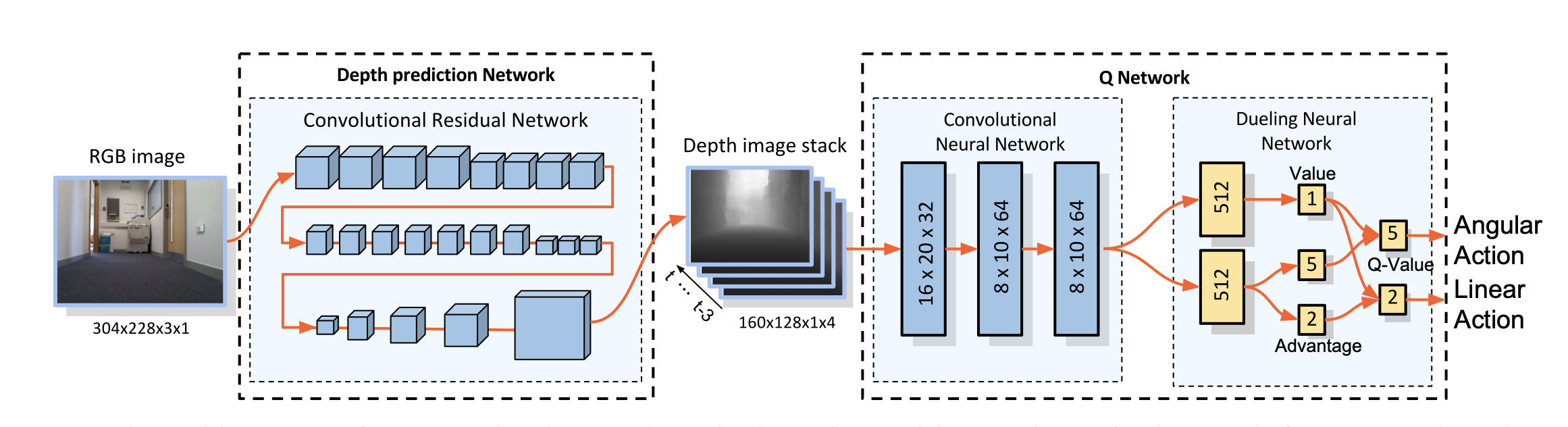}
    \caption{A DRL structure in which an RGB image is fed to a fully  convolutional neural network is firstly constructed and depth is predicted en route towards calculating the actions needed to avoid obstacles \cite{xie2017towards} }
    \label{fig:RLarch}
\end{figure*}

An end-to-end deep reinforcement learning approach for a mobile robot navigating an unknown environment is portrayed by \cite{tai2016towards} in which the inputs are raw depth images and the control commands serve as the outputs. As a means to create a feature representation, they use a CNN architecture with three convolutional  layers which weights are initialized by a pre-trained model  and three fully-connected layers for  exploration  policy  learning. Since the robot is navigating in an indoor environment filled with obstacles, the feedback in terms of negative reward is obtained by the potential collision between the robot and obstacles. After training the model for many thousands of iterations, simulative and real-world indoor experiments demonstrated efficacy in obstacle avoidance along with improvement in the traversable areas detection. A dueling architecture (Figure \ref{fig:RLarch})  named D3QN was initiated in the work of Xie et al. \cite{xie2017towards} 
in the pursuance of obstacle  avoidance using the concepts of convolution and \textit{deep Q Learning} as its main foundations. It consists of a fully convolutional neural network, that outputs depth information from an RGB image which has previously been blended with additional noise and blur to adapt to real-world scenarios, followed by a deep Q network \cite{mnih2013playing} that encloses a convolutional and a dueling network \cite{wang2016dueling} while the main hypothesis implies that the training ,taking place in simulation (Gazebo), will provide adequate knowledge to be transferred to real-world implementations. In terms of training speed, it was proven that the D3QN architecture is almost twice faster than DQN, highlighting its efficiency on obstacle avoidance scenarios and, consequently in efforts of  determining traversable obstacle-free regions. Zhang et al.  \cite{zhang2018robot} 
present an architecture that accepts depth images along with the environment's obtained elevation map and the robot's 3D orientation as the inputs, which are fused and fed into an Advantage Actor-Critic (A3C) model \cite{mnih2016asynchronous}.
 Before their merging, depth and elevation map information is each passed through a four layered convolutional structure followed by a pooling layer whereas the 3D orientation is passed directly to a FC layer and then merged with the elevation information. All input sources are then concatenated and fed to an LSTM that can improve capturing the underlying states of a partially observed environment. Eventually, the actor and critic components consist of an FC layer each with the difference that in the actor part the output vector's values are normalized by the Softmax function. Although both training and testing phases took place in a simulative 3D environment with varying levels of terrain's traversability, their results showed that the agent sufficiently learnt to traverse different terrains towards a predefined goal location, and occasionally around non-traversable objects, with an average Deep RL decision-making time of 0.074 seconds.

An off-policy algorithm \cite{kahn2021badgr} acting as a powerful tool in improving the learning capabilities of an end-to-end RL approach, named BADGR (Berkeley Autonomous Driving Ground Robot), uses a self-supervised data labelling mechanism that is not built upon any human supervision or SLAM techniques in simulation. By collecting data using a random control policy, collisions are detected either by LIDAR or IMU as the robot stores the sensory observations along with the corresponding actions taken. Events are labeled in a self-supervised manner by the collected dataset and then appended to it. Input RGB images act as the current observation that along with a future sequence of actions such as control commands, future events can be predicted. As far as the model's architecture is concerned, the input images are fed into three convolutional and 4 FC layers each one followed(except for the last FC) by a RELU activation function,  in order to form the initial hidden state of a recurrent LSTM unit that will handle each future action and yield the corresponding predicted future event (Figure \ref{fig:badgrarch}). After deploying the BADGR system in real-world environments, it was shown that, by using only 42 hours of autonomously collected data, it could successfully traverse areas of tall vegetation and bumpy terrains.

By taking the history of proprioceptive states into account, Lee et al. \cite{lee2020learning} undertake the rough terrain traversability estimation as a temporal problem that requires a robust controller to produce the appropriate actuation. In this context, a sequential Temporal Convolutional Network (TCN), comprised of convolutional layers followed by a relu activation function, uses input from joint encoders and IMU and, accordingly, implicitly learns to analyze contact and slippage events while a four-legged robot is navigating in complex terrains including those of mud,sand,snow etc. Towards this goal, they claim that direct RL techniques might not be fruitful due to large time processing and thus a teacher-student policy is selected instead. First, the teacher policy based on ground-truth knowledge concerning the interaction between the robot and the terrain, is trained on simulation. Afterwards, it supervises a student learning and the eventual student policy acts on the real robot. An additional concept introduced during the training stage, that enhances the robustness of the method is encapsulated on the adaptive nature of synthesized terrains in order for the controller to traverse them. Finally, as a means to integrate the neural network to regulate the controller, the Policies Modulating Trajectory Generators (PMTG) is employed.

Other methods include the implementation of deep inverse reinforcement learning \cite{wulfmeier2015maximum} for determining off-road traversability \cite{zhu2019off}. Towards this direction, they propose a two-CNN structure that encompasses the vehicle kinematics in the states (2D poses)  which unavoidably leads to an increase of the state-space complexity. In their experiments, data is collected from a laser scanner which, is transformed to input features for a five-layered FCN structure that by recurrently applying a convolution layer for 120 and 150  times per network,  the value  iteration  is  completed  with noticeable reduction of the computational burden. Their results showed  improvements on safe trajectory  prediction.

\begin{table*}[t]
\centering
\renewcommand*{\arraystretch}{1.2}
\caption{Overview of papers that do not rely on supervision (labeled data)} \label{nonSUPERVISEDtable}
\begin{center}
\begin{tabular}{|c|c|c|c|c|c|}
  \hline
  Paper & Input & Data & Labels & Architecture & Platform-specific \\
  \hline\hline
 Lee et al., 2020 \cite{lee2020learning}  & Joint encoders, IMU & 400 steps/episode& RL & CNN,\par MLP & Yes\\\hline
  Tang et al., 2019 \cite{tang2019autonomous}  & Map Cells  &  Unspec & Unsuper&  K-means&  No \\\hline

  Hirose et al., 2019a \cite{hirose2019vunet}   & RGB & 47k & Semi & Encoder-Decoder & No   \\ \hline
  Hirose et al., 2019b  \cite{hirose2019deep} &RGB  & 10.30 hours  & Semi & Encoder-Decoder  & No \\
  & & & &MPC  &  \\\hline

    Zhu et al., 2019   \cite{zhu2019off} & Lidar feature map & 2.4k scene maps & RL & CNN &  No \\ \hline 
    Hirose et al., 2018  \cite{hirose2018gonet}  &  RGB & 78k & Semi  & GAN & No   \\ \hline
  Zhang et al.,2018a  \cite{zhang2018long}  & Depth  Orientation & 208 steps/episode & RL & CNN & No \\
  & Elevation maps & &  & LSTM &No \\ \hline

  Xie et al., 2017 \cite{xie2017towards} & RGB, Depth & 500 steps/episode & RL & CNN & No\\\hline
  Suger et al., 2015  \cite{suger2015traversability} & 3D- Laser & Unspec & Semi & Naive-Bayes &   No\\ \hline 

Shneier et al.,2008 \cite{shneier2008learning}  & Stereo & Unspec & Unsuper & Occupancy maps, &Yes \\ 
  & & & & Geometric Vision &  \\ \hline 
Kim et al., 2006  \cite{kim2006traversability}  & Stereo & 220k patches & Non-trainable & Geometric Vision  & No  \\ \hline

\end{tabular}
\end{center}
\end{table*}

\section{Research Challenges}

\begin{figure}[t]
    \centering
    \includegraphics[width=8.5cm]{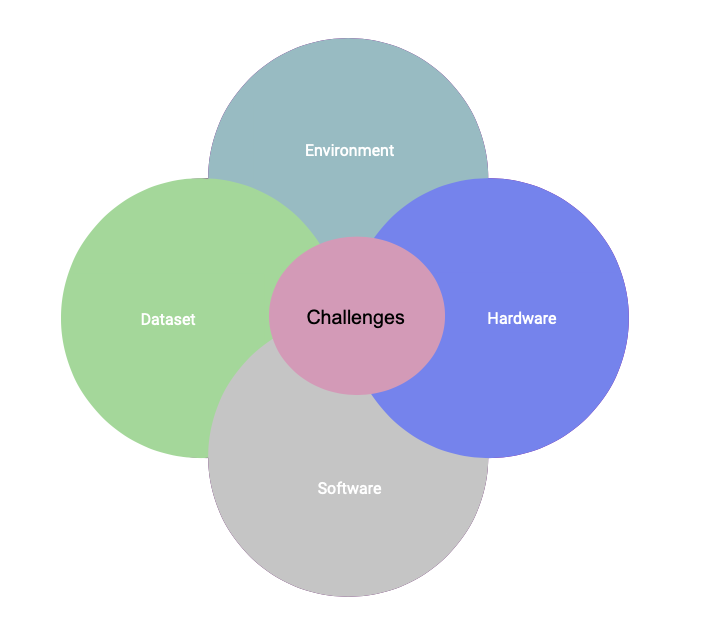}
    \caption{An interconnection among the different type of challenges occurring in traversability estimation scenarios}
    \label{fig:kch}
\end{figure}

Natural outdoor terrains impose challenges due to the unique structure that 
each environment exhibits. For this reason the challenges enforced can be observed directly (uneven surfaces, terrain singularities etc.) while others can be substantially treacherous for the robot's safety particularly in the cases of non-compliant objects that appear to be traversable, muddy terrains and undubitably, erratic shifts in lighting and weather conditions. 
On the other hand, ML and especially DL techniques require an extensive amount of data, a fact that frequently poses stringent questions to the available computational power that handles this data. Before that, another challenge regarding the successful collection of meaningful data arises. Therefore, careful engineering of the hardware integrated on the platform as well as precision of the algorithm responsible for accurate data gathering and labelling, for instance in SSL, become two main factors of arduousness encountered in traversability experiments. What is more, dataset construction along with meticulously deploying the software (simulation/algorithms) to the real-world platform are two additional topics that require precision and robustness. Hence, manual labelling would eventually require a skilled human expert who can guide the agent in areas that terrain information is nebulous.

\begin{table*}[t]
\centering
\renewcommand*{\arraystretch}{1.2}
\caption{Dataset/Software/Hardware-related Challenges encountered in ML/DL references }

\begin{center}
\begin{tabular}[t]{{|c|c|c|c}}
  \hline
 Challenges & Task & References \\
  \hline\hline

Computational Complexity    & Semantic Terrain Segmentation& \cite{gao2021fine} \cite{goh2022mars}\\
  & Terrain Exploration & \cite{zhang2018robot}\\
  & Terrain Classification &   \cite{goh2022mars}
    \\

    &  Scene Traversability& \cite{hirose2017go}   \\

 \hline

 Dataset Construction 
&Terrain Exploration&  \cite{sekar2020planning} \cite{zhang2018robot}\\
 & Terrain Classification &  \cite{narvaez2018terrain} \cite{brooks2007self} \\

 & Semantic Terrain Segmentation &  \cite{zurn2020self} \\
 & Navigation &  \cite{kahn2021badgr} \\
 & Collision prediction model &  \cite{richter2017safe} \\\hline

Obstacle Avoidance  

& Terrain Traversability Analysis &  \cite{ho2013traversability} \cite{suger2015traversability}\\
& Indoor Exploration & \cite{tai2017autonomous}   \cite{tang2019autonomous} \\

& Generalization & \cite{dima2004classifier} \cite{palazzo2020domain} \\
& Motion Planning & \cite{shah2022viking}\\
& Route Traversability Prediction & \cite{palazzo2020domain}\\
& Terrain Cost Calculation&\cite{ollis2007bayesian}  \\
\hline

Localization Uncertainty &  Mapping Algorithm& \cite{fankhauser2018probabilistic} \\

& Terrain Traversability Analysis & \cite{oliveira2020bayesian} \\
\hline

 Robust Slip Prediction model & Terrain  Classification & \cite{rothrock2016spoc} \\\hline

Sensory Information Extraction & Motion Planning  &\cite{pfeiffer2017perception} \\\hline

Simulation to Real world deployment & Collision Avoidance & \cite{xie2017towards} \cite{zhang2018long}\\
&Motion Planning &  \cite{pfeiffer2017perception} \\\hline

\end{tabular}

\end{center}
\label{tab: datachal}
\end{table*}

Therefore, we can state that
the challenges traversability estimation scenarios can be a result of environment's complexity, dataset construction as well as software/hardware-related issues. It becomes apparent that these aforementioned challenges are interconnected as, for instance, a complex and unpredictable environment directly affects the dataset collection as well as the accuracy of the software representation (Figure \ref{fig:kch}).

\begin{figure}[t]
    \centering
    \includegraphics[width=8.5cm]{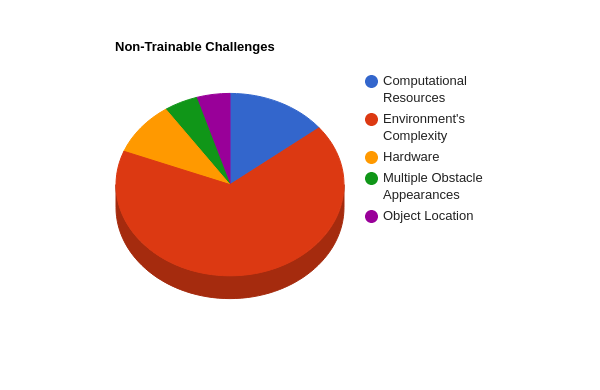}
    \caption{Challenges in non-trainable methods}
    \label{fig:Challengesnon}
\end{figure}

\begin{figure}[t]
    \centering
    \includegraphics[width=8.5cm]{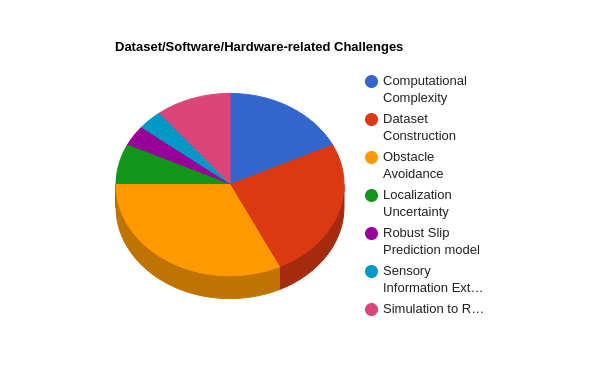}
    \caption{Dataset/Software/Hardware Challenges in ML,DL methods}
    \label{fig:Challengesdsw}
\end{figure}

\begin{figure}[t]
    \centering
    \includegraphics[width=8.5 cm]{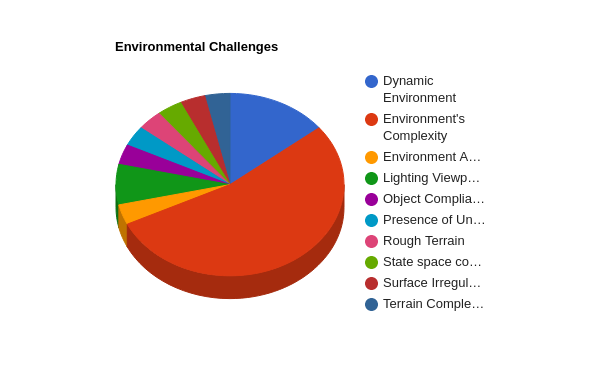}
    \caption{Environmental Challenges in ML,DL methods}
    \label{fig:Challengesenb}
\end{figure}

Irrespective of the technique used to infer traversability, these challenges will be a perpetual factor to consider  and address in traversabiity estimation problems as a way to reinforce and secure the platform's safety as well as the fruitful accomplishment by the experiment. Tables  \ref{tab: datachal}, \ref{tab: Envchal},  \ref{tab: nonmlchal} give an overview of the challenges encountered by the researchers while  using different types of learning methods for outdoor (and merely indoor)  traversability-related tasks. After quantifying some information extracted from the Tables  \ref{tab: datachal},\ref{tab: Envchal},\ref{tab: nonmlchal}, we observe that in Figures \ref{fig:Challengesnon}, \ref{fig:Challengesdsw} and \ref{fig:Challengesenb}, environment-related challenges has been consistently present in traversability estimation problems.

\begin{table*}[t]
\centering
\renewcommand*{\arraystretch}{1.2}
\caption{Environmental Challenges encountered in ML/DL references }

\begin{center}
\begin{tabular}[t]{{|c|c|c|c}}
  \hline
 Challenges & Task & References \\
  \hline\hline

 Dynamic Environment & Terrain Traversability Analysis& \cite{sock2016probabilistic}\\
& Semantic Segmentation&  \cite{valada2017adapnet}\\
& Learning-based Navigation System &  \cite{shah2021ving} \\
 & Terrain  Classification 
 & \cite{kingry2018vision} \\ \hline

\hline

   Environment's Complexity&Traversability Classification &    
 \cite{happold2006enhancing}  \cite{kim2007traversability} \cite{sikand2021visual}  \\

     & Indoor Navigation & \cite{hirose2018gonet} \cite{hirose2019vunet} \cite{hirose2019deep} \\ 
  
  & Terrain Classification &   \cite{brooks2007self} \cite{zhou2012self}
    \\
    
 & Outdoor/Off-road Navigation &  \cite{zhang2018long} \cite{reina2012towards} \\
& Mapping System& \cite{droeschel2017continuous}\\
 &Terrain Traversability Analysis & \cite{suger2015traversability}    \\
 
& Semantic Terrain Segmentation & \cite{zurn2020self}\\

 & Learning-based Exploration &  \cite{tai2016towards} \\
  & Motion Planning & \cite{belter2016adaptive} \\

 \hline
    Lighting,Viewpoint Invariancy/Consistency & Terrain Classification &  \cite{hadsell2009learning}  \cite{rothrock2016spoc} \\\hline
Environment Adaptation & Collision Avoidance &\cite{xie2017towards} \\\hline

 Object Compliancy & Navigation & \cite{kahn2021badgr} \\\hline  
Presence of Unknown Obstacles & Traversability Classification,Anomaly Detection & \cite{wellhausen2019should} \\\hline

Rough Terrain   &  Locomotion Control & \cite{lee2020learning} \\\hline

 State-space Complexity & Terrain Traversability Analysis &   \cite{zhu2019off} \\\hline

Surface Irregularity  &Road Traversability Detection & \cite{bellone2017learning} \\ \hline

Terrain Complexity  & Traversability  Estimation Framework  & \cite{chavez2018learning}   \\\hline
\hline  
\end{tabular}

\end{center}
\label{tab: Envchal}
\end{table*}

\begin{table*}[t]
\centering
\renewcommand*{\arraystretch}{1.2}
\caption{Challenges encountered in papers with non-trainable methods} 
\label{nonmlchall}
\begin{center}
\begin{tabular}{|c|c|c|}
  \hline
Challenges & Task & References \\
  \hline\hline

Computational Resources & Outdoor/Off-road Navigation &  \cite{kim2020vision}\cite{langer1994behavior} \\
& Motion Planning& \cite{pan2019gpu}  \\
\hline

  Environment's Complexity&Terrain Traversability Analysis &  \par \cite{bajracharya2013high}\cite{shneier2008learning} \cite{fan2021step}   \cite{li2019rugged}  \\
    & &  \cite{bogoslavskyi2013efficient} \cite{loc2011improving}  \cite{ye2007navigating}\cite{castejon2005traversable} \\
  & Outdoor/Off-road Navigation &  \cite{huertas2005stereo}\cite{kim2020vision}\cite{langer1994behavior} \\
 
 & Traversability Classification &   \cite{kim2006traversability}  
    \\

& Traversability Index Definition& \cite{seraji1999traversability}
 \\
 & Traversability Map & \cite{martin2013building}\\
 
 \hline

Hardware-Related & Outdoor/Off-road Navigation &  \cite{kim2020vision}\\
& Path Planning& \cite{ishigami2011path}\\

\hline  

  Multiple Obstacle  Appearances   & Outdoor  Navigation &  \cite{rankin2005evaluation} \\\hline 

  Object Location   &Object  Detection   & \cite{ulrich2000appearance} \\\hline

\end{tabular}
\end{center}
\label{tab: nonmlchal}
\end{table*}

\space \space \space \space \space \space \space \space \space 

\section{Conclusions}

An abundant number of robotic applications in both outdoor and indoor environments constitutes the concept of traversability estimation a central task for safe and successful navigation through motion planning. As different environments are governed by various levels of stochasticity, the efforts to collect and interpret data from various sensor modalities often reveal further challenges due to the type and the volume of the data acquired. In this survey, we examine how the acquisition of sensory data, in various trainable and non-trainable methods, has been interpreted towards deriving fruitful conclusions for outdoor and indoor traversability estimation experiments.

Traditional geometric methods representing the environment in 3D have been used in conjunction with bespoke algorithms  to identify obstacle and relevant pixels in the scenes encountered. Frequently witnessed, early traversability estimation computer vision techniques used certain assumptions in order to make predictions based on the outputs of obstacle detection algorithms. This led to limitations regarding long-distance perception,deformable ground traversability and generally speaking, scientific approaches were more conservative in their implementations.
Afterwards, conventional machine learning methods examined the use of merging multiple sensor modalities along with probabilistic approaches that can model environment's uncertainty. However,features' engineering is difficult and time-consuming for the human expert. Due to the progression of the algorithmic learning techniques, different types of learning are associated with different types of encountered challenges.

Extracting meaningful conclusions in relation to the traversability of the scene has recently been associated, on the grounds that deep learning techniques are widely popular nowadays, with the magnitude of collected data. Powerful deep learning algorithms often require enormous datasets in order to adequately train the weights of the model making headway for multiple downstream tasks, such as image classification, semantic segmentation etc. By the same token, due to the difficulty to create models that demonstrate high-level representations, the need to generate accurate labels about the terrain or the surrounding obstacles, while addressing the various challenges arising during data collection, is closely correlated to the type of the learning technique that is implemented. This survey documents an assortment of preceding and present-day learning techniques whilst it distinguishes the impact of employing Self-Supervised Learning techniques, in contemporary traversability estimation tasks. 
In proximity to the Self-Supervised Learning techniques, contemporary and future research endeavors target the use of transfer learning or representation-learning methods as a propitious way to build data-efficient and robust traversability learning frameworks while truncating the highly cumbersome and vigilant tasks of manual labelling by a human expert.






\bibliographystyle{unsrt}  
\bibliography{references}

\end{document}